\documentclass{article}
\usepackage{ijcai23}

\usepackage{times}
\usepackage{soul}
\usepackage{url}
\usepackage[hidelinks]{hyperref}
\usepackage[utf8]{inputenc}
\usepackage[small]{caption}
\usepackage{graphicx}
\usepackage{amsmath}
\usepackage{amsthm}
\usepackage{booktabs}
\usepackage{algorithm}
\usepackage{algorithmic}
\usepackage[switch]{lineno}
\usepackage{multirow}
\usepackage[table,xcdraw]{xcolor}
\definecolor{myurlcolor}{rgb}{0,0.5,0.5}  
\usepackage{tabu}
\newcommand{\xz}[1]{\textcolor{black}{#1}}

\hypersetup{
  colorlinks=true,  
  urlcolor=myurlcolor,  
  citecolor=black,  
  linkcolor=black,  
}

\title{Large Language Model based Multi-Agents: A Survey of Progress and Challenges}

\author{
Taicheng Guo$^1$
\and
Xiuying Chen$^2$\and
Yaqi Wang$^3$\thanks{This work was done when Yaqi was visiting students at the University of Notre Dame.} \and
Ruidi Chang \and 
Shichao Pei$^4$ \and \\
Nitesh V. Chawla$^1$ \and
Olaf Wiest$^1$ \and
Xiangliang Zhang$^1$\thanks{Corresponding author.}
\affiliations
$^1$University of Notre Dame, $^2$King Abdullah University of Science and Technology\\
$^3$Southern University of Science and Technology,
$^4$University of Massachusetts Boston
\emails
\{tguo2, nchawla, owiest, xzhang33\}@nd.edu,
xiuying.chen@kaust.edu.sa,
ywang84@nd.edu, ruidic@alumni.cmu.edu, shichao.pei@umb.edu
}

\begin{document}

\maketitle

\begin{abstract}
Large Language Models (LLMs) have achieved remarkable success across a wide array of tasks. Due to the impressive planning and reasoning abilities of LLMs, they have been used as autonomous agents to do many tasks automatically. Recently, based on the development of using one LLM as a single planning or decision-making agent, LLM-based multi-agent systems have achieved considerable progress in complex problem-solving and world simulation. 
To provide the community with an overview of this dynamic field,  we present this survey to offer an in-depth discussion on the essential aspects of multi-agent systems based on LLMs, as well as the challenges. Our goal is for readers to gain substantial insights on the following questions: What domains and environments do LLM-based multi-agents simulate? How are these agents profiled and how do they communicate? What mechanisms contribute to the growth of agents' capacities? For those interested in delving into this field of study, we also summarize the commonly used datasets or benchmarks for them to have convenient access. To keep researchers updated on the latest studies,  we maintain an open-source \href{https://github.com/taichengguo/LLM_MultiAgents_Survey_Papers}{GitHub repository}, dedicated to outlining the research on LLM-based multi-agent systems. 

\end{abstract}

\section{Introduction}


Large Language Models (LLMs) have recently shown remarkable potential in reaching a level of reasoning and planning capabilities comparable to humans.
This ability exactly aligns with the expectations of humans for autonomous agents that can perceive the surroundings, make decisions, and take actions in response~\cite{xi2023rise,Wooldridge1995IntelligentAT,Artificial_Intelligence_AModernApproach,guo2023indeed,liang2023let}. Hence, LLM-based agent has been studied and rapidly developed to understand and generate human-like instructions, facilitating sophisticated interactions and decision-making in a wide range of contexts
~\cite{yao2023tree,shinn2023reflexion,li2023apibank}. Timely survey papers systematically summarize the progress of LLM-based agents, as seen in works ~\cite{xi2023rise,wang2023survey}.


\begin{figure*}[ht]
\begin{center}
\includegraphics[scale=0.56]{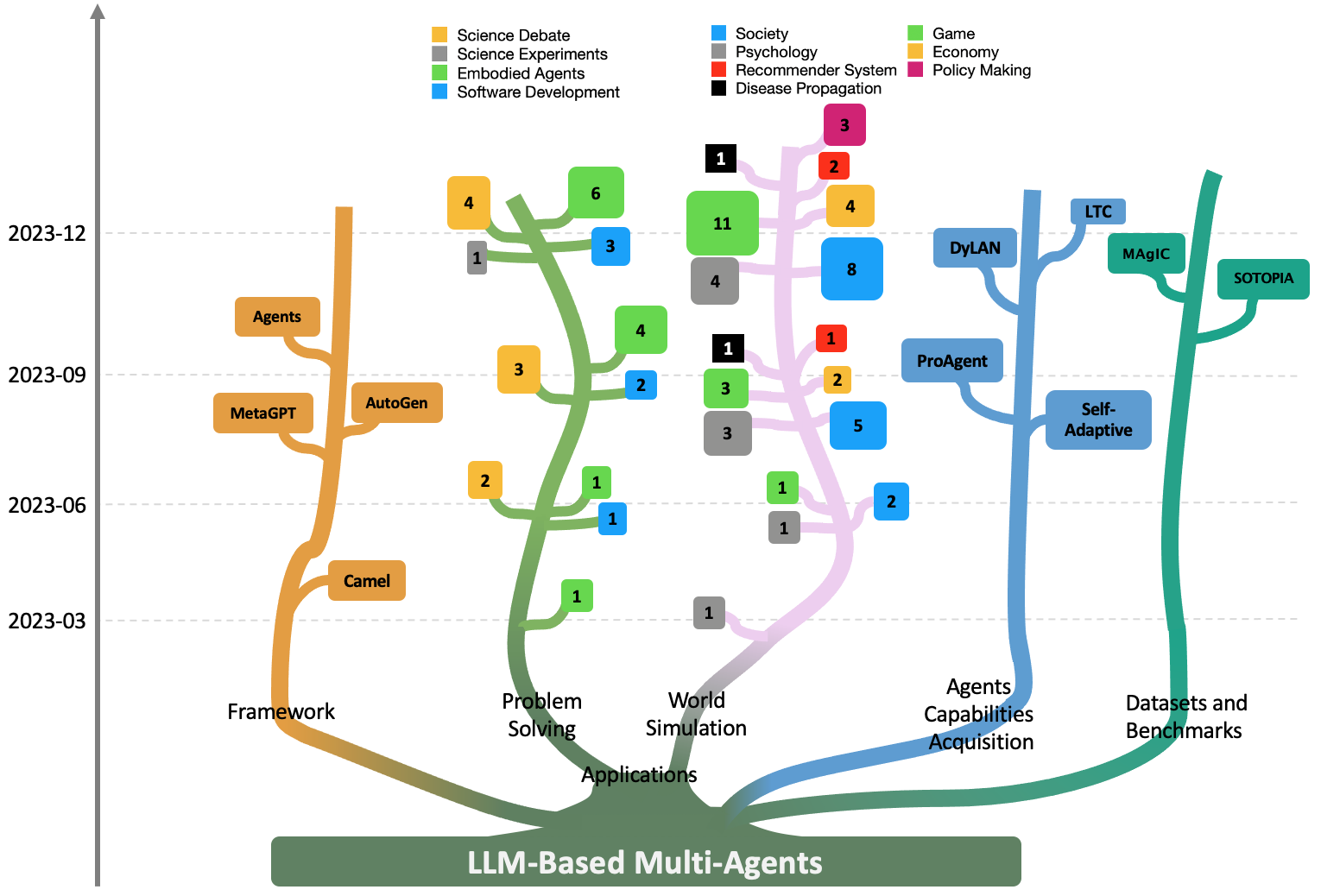}
\end{center}
\caption{The rising trend in the research field of LLM-based Multi-Agents. For Problem Solving and World Simulation, we categorize current work into several categories and count the number of papers of different types at 3-month intervals. The number at each leaf node denotes the count of papers within that category.}
\label{trend}
\end{figure*}


Based on the inspiring capabilities of the single LLM-based agent, LLM-based Multi-Agents have been proposed to leverage the collective intelligence and specialized profiles and skills of multiple agents. 
\xz{Compared to systems using a single LLM-powered agent, multi-agent systems offer advanced capabilities by \underline{1)} specializing LLMs into various distinct agents, each with different capabilities, and \underline{2)} enabling interactions among these diverse agents to simulate complex real-world environments effectively. In this context, multiple autonomous agents collaboratively engage in planning, discussions, and decision-making, mirroring the cooperative nature of human group work in problem-solving tasks. This approach capitalizes on the communicative capabilities of LLMs, leveraging their ability to generate text for communication and respond to textual inputs. Furthermore, it exploits LLMs' extensive knowledge across various domains and their latent potential to specialize in specific tasks. Recent research has demonstrated promising results in utilizing LLM-based multi-agents for solving various tasks,}
such as software development~\cite{hong2023metagpt,qian2023communicative}, multi-robot systems~\cite{mandi2023roco,zhang2023building}, 
society simulation~\cite{park2023generative,park2022social}, policy simulation~\cite{xiao2023simulating,hua2023war}, and game simulation~\cite{xu2023language,wang2023avalon}. 
\xz{Due to the nature of interdisciplinary study in this field, it has attracted a diverse range of researchers, expanding beyond AI experts to include those from social science, psychology, and policy research. The volume of research papers is rapidly increasing,}
as shown in Fig. \ref{trend} (inspired by the design in ~\cite{gao2023retrieval}),
thus broadening the impact of LLM-based Multi-Agent research.
Nonetheless, earlier efforts were undertaken independently,  resulting in an absence of a systematic review to summarize them, establish comprehensive \xz{blueprint of this field}, and examine future research challenges. This underscores the significance of our work and serves as the motivation behind presenting this survey paper, dedicated to the research on LLM-based multi-agent systems.


We expect that our survey can make significant contributions to both the research and development of LLMs and to a wider range of interdisciplinary studies employing LLMs. Readers will gain a comprehensive overview of  LLM-based Multi-Agent (LLM-MA) systems, grasp the fundamental concepts involved in establishing multi-agent systems based on LLMs, and catch the latest research trends and applications in this dynamic field. We recognize that this field is in its early stages and is rapidly evolving with fresh methodologies and applications. To provide a sustainable resource complementing our survey paper, we maintain an open-source GitHub repository\footnote{\scriptsize \url{https://github.com/taichengguo/LLM_MultiAgents_Survey_Papers}}.
We hope that our survey will inspire further exploration and innovation in this field,
as well as applications across a wide array of research disciplines.

To assist individuals from various backgrounds in understanding LLM-MA techniques and to complement existing surveys by tackling unresolved questions, we have organized our survey paper in the following manner. After laying out the background knowledge in Section~\ref{Sec2:background}, we address a pivotal question: \emph{How are LLM-MA systems aligned with the collaborative task-solving environment}?   To answer this, we present a comprehensive schema for positioning, differentiating, and connecting various aspects of LLM-MA systems in  \textbf{Section~\ref{Sec3:Taxonomy}}. We delve into this question by discussing: 1) the  \textbf{agents-environment interface}, which details how agents interact with the task environment; 2)  \textbf{agent profiling}, which explains how an agent is characterized by an LLM to behave in specific ways; 3)  \textbf{agent communication}, which examines how agents exchange messages and collaborate; and 4)  \textbf{agent capability acquisition}, which explores how agents develop their abilities to effectively solve problems. 
An additional perspective for reviewing studies about LLM-MA is their application. In  \textbf{Section~\ref{Sec5:Application}}, we categorize current  \textbf{applications} into two primary streams: multi-agents for \textbf{problem-solving} and multi-agents for \textbf{world simulation}.
To guide individuals in identifying appropriate \textbf{tools and resources},  we present open-source implementation frameworks for studying LLM-MA, as well as the usable datasets and benchmarks in  \textbf{Section~\ref{Sec4:Framework}}.
Based on the previous summary, we open the discussion for future research challenges and opportunities in Section~\ref{Sec7:Challenges and Opportunities}. The conclusions are summarized in Section~\ref{Sec8:Conclusion}.
\section{Background}
\label{Sec2:background}
\subsection{Single-Agent Systems Powered LLMs}
We introduce the background by first outlining the capabilities of a single-agent system based on  LLMs, following the discussion presented in \cite{agent_link}. 

\paragraph{Decision-making Thought:} 
This term denotes the capability of LLM-based agents, guided by prompts, to break down complex tasks into smaller subgoals~\cite{khot2023decomposed}, think through each part methodically (sometimes exploring multiple paths)~\cite{yao2023tree}, and learn from past experiences~\cite{shinn2023reflexion} to perform better decision-making on complex tasks. This capability enhances the autonomy of a single LLM-based agent and bolsters its effectiveness in problem-solving.

\paragraph{Tool-use:} LLM-based agents' tool-use capability allows them to leverage external tools and resources to accomplish tasks, enhancing their functional capabilities and operate more effectively in diverse and dynamic environments~\cite{li2023apibank,ruan2023tptu,gao2023retrieval}.  

\paragraph{Memory:} This ability refers to the capability of LLM-based agent for conducting in-context learning~\cite{dong2023survey} as short memory or external vector database~\cite{lewis2021retrievalaugmented} as long memory to preserve and retrieve information over prolonged periods~\cite{wang2023survey}. This ability enables a single LLM-based agent to maintain contextual coherence and enhance learning from interactions. 

\subsection{Single-Agent VS. Multi-Agent Systems}
Single-Agent systems empowered by LLMs have shown inspiring cognitive abilities \cite{sumers2023cognitive}. 
The construction of such systems concentrates on formulating their internal mechanisms and interactions with the external environment. Conversely, 
LLM-MA  systems emphasize diverse agent profiles, inter-agent interactions, and collective decision-making processes. From this perspective, more dynamic and complex tasks can be tackled by the collaboration of multiple autonomous agents, each of which is equipped with unique strategies and behaviors, and engaged in communication with one another.

\section{Dissecting LLM-MA Systems: Interface, Profiling, Communication, and Capabilities}
\label{Sec3:Taxonomy}
\begin{figure*}[ht]
\centering
\includegraphics[width= 0.73\textwidth]{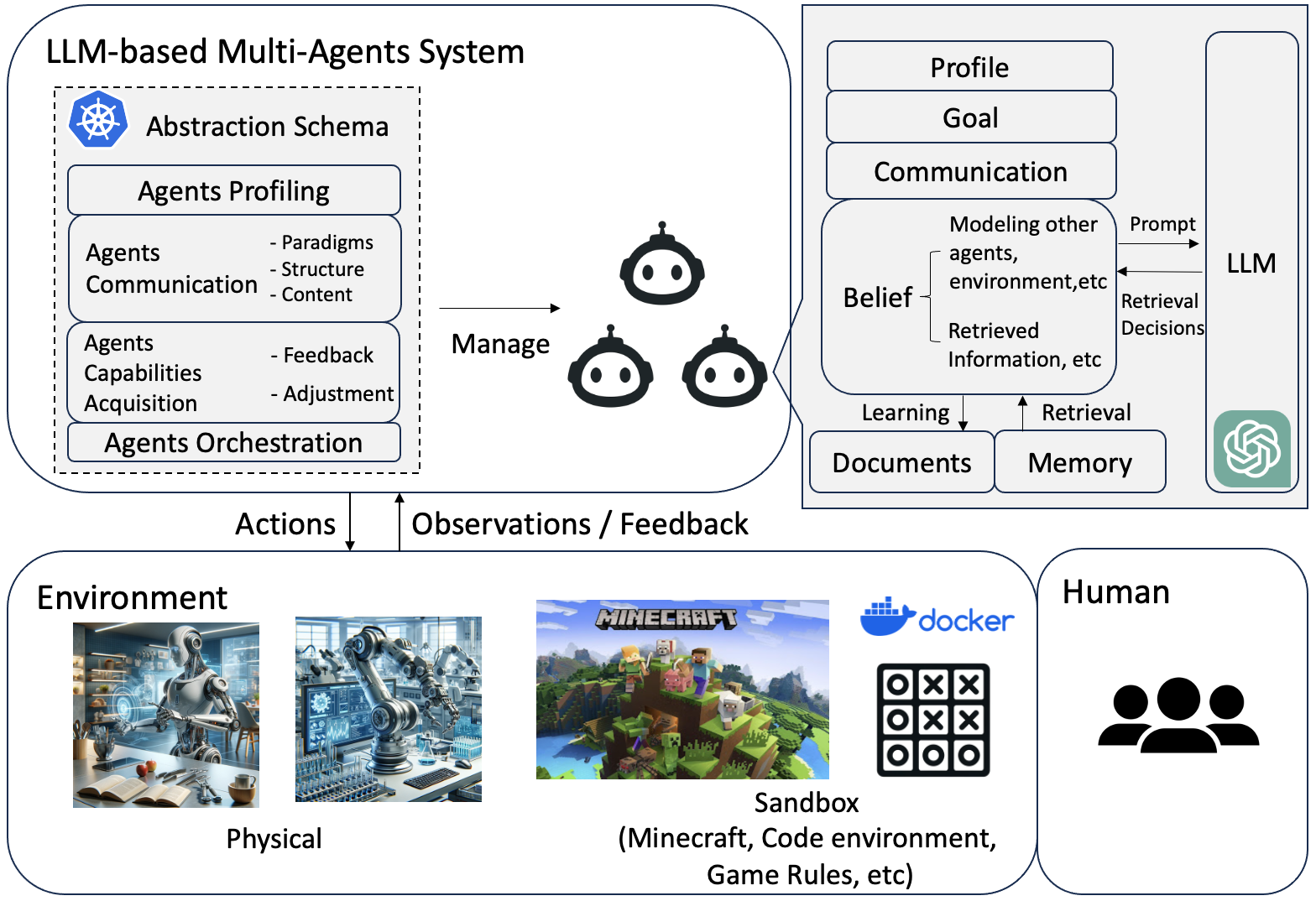}
\caption{The Architecture of LLM-MA Systems.} 
\label{fig:Architecture} 
\end{figure*}

In this section, we delve into the intricacies of LLM-MA systems, where multiple autonomous agents engage in collaborative activities akin to human group dynamics in problem-solving scenarios. \xz{A critical inquiry we address is how these LLM-MA systems are  aligned  to  their operational environments and the collective objectives they are designed to achieve. To shed light on this, we present the general architecture of these systems  in Fig. \ref{fig:Architecture}.} Our analysis dissects the operational framework of these systems, focusing on four key aspects: the agents-environment interface, agent profiling, agent communication, and agent capability acquisition.

\subsection{Agents-Environment Interface}
\xz{The operational environments defines the specific contexts or settings in which the LLM-MA systems are deployed and interact. 
For example, these environments can be like software development \cite{hong2023metagpt}, gaming \cite{mao2023alympics}, and various other domains such as financial markets \cite{li2023tradinggpt}  or even social behavior modeling \cite{park2023generative}.  The LLM-based agents perceive and act within the environment, which in turn influences their behavior and decision making. For example, in the Werewolf Game simulation, the sandbox environment sets the game's framework, including transitions from day to night, discussion periods, voting mechanics, and reward rules. Agents, such as werewolves and the Seer, perform specific actions like killing or checking roles. Following these actions, agents receive feedback from the environment, informing them of the game's current state. This information guides the agents in adjusting their strategies over time, responding to the evolving gameplay and interactions with other agents.}
The Agents-Environment Interface refers to the  way in which agents interact with and perceive the environment. It's through this interface that agents understand their surroundings, make decisions, and learn from the outcomes of their actions.
We categorize the current interfaces in LLM-MA systems into three types, \emph{Sandbox, Physcial}, and \emph{None}, as detailed in Table \ref{summary}.
The \emph{\textbf{Sandbox}} refers to a simulated or virtual environment built by human where agents can interact more freely and experiment with various actions and strategies. This kind of interface is widely used in software development (code interpreter as simulated environment)~\cite{hong2023metagpt}, gaming (using game rules as simulated environment)~\cite{mao2023alympics}, etc.  The \emph{\textbf{Physical}} is a real-world environment where agents interact with physical entities and obey real-world physics and constraints. In physical space, agents normally need to take actions that can have direct physical outcomes. For example, in tasks such as sweeping the floor, making sandwiches, packing groceries, and arranging cabinets, robotic agents are required to perform actions iteratively, observe the physical environment, and continuously refine their actions~\cite{mandi2023roco}. Lastly, \emph{\textbf{None}} refers to scenarios where there is no specific external environment, and agents do not interact with any environment. For example, many applications~\cite{du2023improving,xiong2023examining,chan2023chateval} utilize multiple agents to debate a question to reach a consensus. These applications primarily focus on communication among agents and do not depend on the external environment.

\subsection{Agents Profiling}

In LLM-MA systems, agents are defined by their traits, actions, and skills, which are tailored to meet specific goals. Across various systems, agents assume distinct roles, each with comprehensive descriptions encompassing characteristics, capabilities, behaviors, and constraints. For instance, in gaming environments, agents might be profiled as players with varying roles and skills, each contributing differently to the game's objectives. In software development, agents could take on the roles of product managers and engineers, each with responsibilities and expertise that guide the development process. Similarly, in a debating platform, agents might be designated as proponents, opponents, or judges, each with unique functions and strategies to fulfill their roles effectively. These profiles are crucial for defining the agents' interactions and effectiveness within their respective environments. Table \ref{summary} lists the agent \emph{\textbf{Profiles}} in recent LLM-MA works.


Regarding the \emph{\textbf{Agent Profiling Methods}}, we 
categorized them into three types: \emph{Pre-defined, Model-Generated}, and \emph{Data-Derived}.
In the \emph{\textbf{Pre-defined}} cases,  agent profiles are explicitly defined by the system designers. 
The \emph{\textbf{Model-Generated}} method creates agent profiles by models, e.g., large language models. The \emph{\textbf{Data-Derived}} method involves constructing agent profiles based on pre-existing datasets.

\subsection{Agents Communication} \vspace{-0.05cm}

The communication between agents in LLM-MA systems is the critical infrastructure supporting collective intelligence. We dissect agent communication from three perspectives: 1) \emph{Communication Paradigms}:   the styles and methods of interaction between agents; 2) \emph{Communication Structure}: the organization and architecture of communication networks within the multi-agent system; and 3) \emph{Communication Content}  exchanged between agents.

\paragraph{Communication Paradigms:} 
Current LLM-MA systems mainly take three paradigms for communication: 
\emph{Cooperative}, \emph{Debate}, and \emph{Competitive}.  \emph{\textbf{Cooperative}}  agents work together towards a shared goal or objectives, typically exchanging information to enhance a collective solution. The \emph{\textbf{Debate}} paradigm is employed when agents engage in argumentative interactions, presenting and defending their own viewpoints or solutions, and critiquing those of others. This  paradigm is ideal for reaching a consensus or a more refined solution. \emph{\textbf{Competitive}} agents work towards their own goals that might be in conflict with the goals of other agents.

\begin{figure}[t]
\centering
\includegraphics[width= 0.44\textwidth]{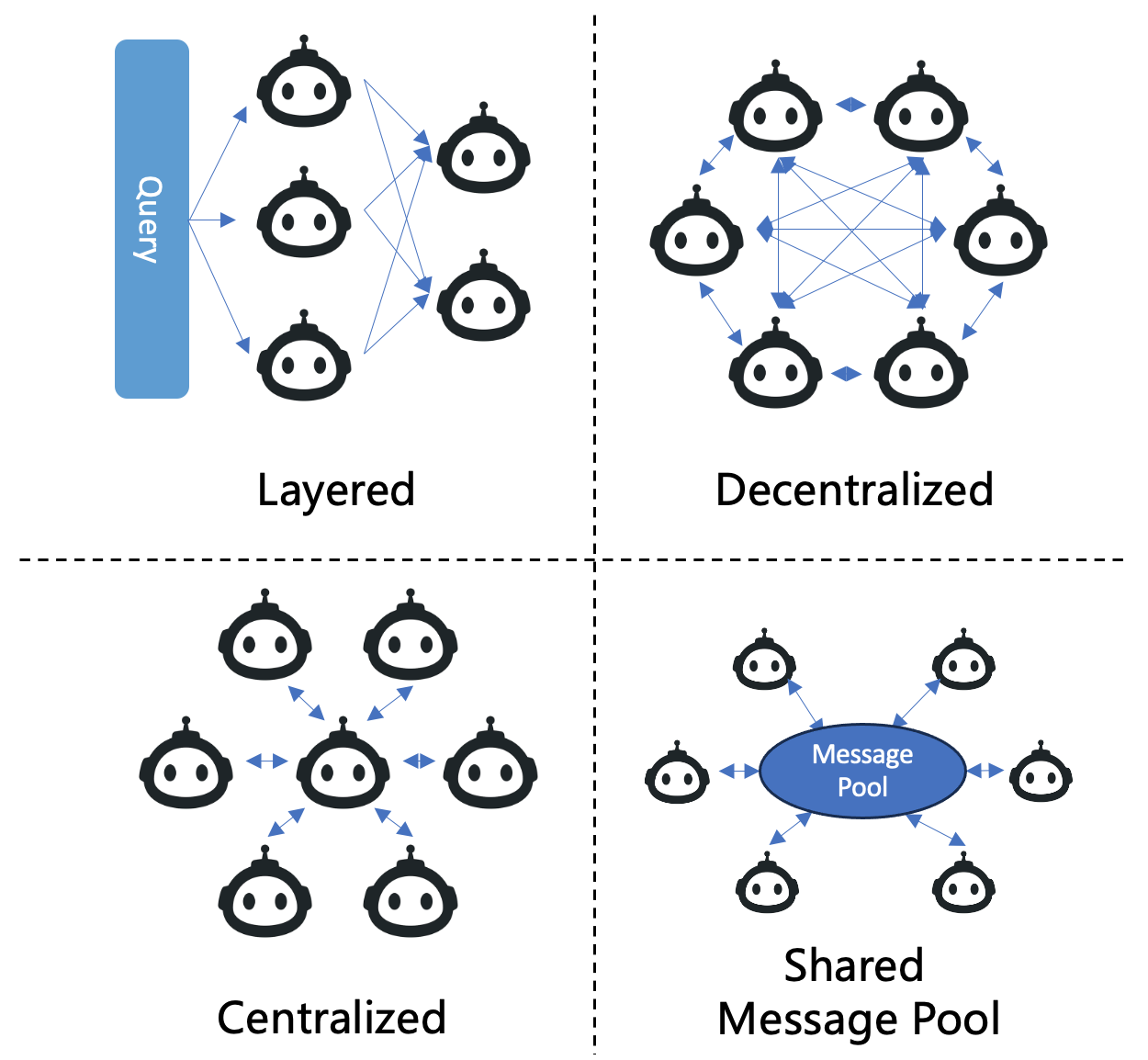}
\caption{The Agent Communication Structure.} 
\label{fig: Agent_communication} 
\end{figure}

\paragraph{Communication Structure:} 
Fig.~\ref{fig: Agent_communication} shows {four} typical communication structures in LLM-MA systems. 
\emph{\textbf{Layered}} communication is structured hierarchically, with agents at each level having distinct roles and primarily interacting within their layer or with adjacent layers. \cite{liu2023dynamic} introduces a framework called Dynamic LLM-Agent Network (DyLAN), which organizes agents in a multi-layered feed-forward network. This setup facilitates dynamic interactions, incorporating features like inference-time agent selection and an early-stopping mechanism, which collectively enhance the efficiency of cooperation among agents. 
\emph{\textbf{Decentralized}} communication operates on a peer-to-peer network, where agents directly communicate with each other, a structure commonly employed in world simulation applications.
\emph{\textbf{Centralized}} communication involves a central agent or a group of central agents coordinating the system's communication, with other agents primarily interacting through this central node.
{\emph{\textbf{Shared Message Pool}} is proposed by MetaGPT~\cite{hong2023metagpt} to improve the communication efficiency. This communication structure maintains a shared message pool where agents publish messages and subscribe to relevant messages based on their profiles, thereby boosting communication efficiency.}

\paragraph{Communication Content:} 
In LLM-MA systems, the Communication Content typically takes the form of text. The specific content varies widely and  depends on the particular application. For example, in software development, agents may communicate with each other about code segments. In simulations of games like Werewolf,  agents might discuss their analyses, suspicions, or strategies.  


\subsection{Agents Capabilities Acquisition}
The Agents Capabilities Acquisition is a crucial process in LLM-MA, enabling agents to learn and evolve dynamically. In this context, there are two fundamental concepts: the types of \emph{\textbf{feedback}} from which agents should learn to enhance their capabilities,  and the \emph{\textbf{strategies}} for agents to \emph{\textbf{adjust}} themselves to effectively solve complex problems. 
\paragraph{Feedback:} Feedback involves the critical information that agents receive about the outcome of their actions, helping the agents learn the potential impact of their actions and adapt to complex and dynamic problems. In most studies, the format of feedback provided to agents is textual. Based on the sources from which agents receive this feedback, it can be categorized into four types. \textbf{1) Feedback from Environment}, e.g.,  from either real world environments or   virtual environments~\cite{wang2023survey}. It is prevalent in most LLM-MA for problem-solving scenarios, including Software Development (agents obtain feedback from Code Interpreter), and Embodied multi-agents systems (robots obtain feedback from real-world or Simulated environments). \textbf{2) Feedback from Agents Interactions}  means that the feedback comes from the judgement of other agents or from agents communications. It is common in problem-solving scenarios like science debates, where agents learn to  critically evaluate and refine the conclusions through communications. In  world simulation scenarios such as Game Simulation,  agents learn to refine strategies based on previous interactions between other agents. \textbf{3)  Human Feedback} comes directly from humans and is crucial for aligning the multi-agent system with human values and preferences.   This kind of feedback is widely used in most ``Human-in-the-loop'' applications~\cite{wang2021putting}. Last \textbf{4) None}.  In some cases, there is no feedback provided to  the agents. This  often happens for world simulation works focused on analyzing simulated results rather than the planning capabilities of agents. In such scenarios, like propagation simulation, the emphasis is on result analysis, and hence, feedback is not a component of the system. 

\paragraph{Agents Adjustment to Complex Problems:} 
To enhance their capabilities, agents in LLM-MA systems can adapt through three main solutions.
\textbf{1) Memory.} Most LLM-MA systems leverage a memory module for agents to adjust their behavior. Agents store information from previous interactions and feedback in their memory. When performing actions, they can retrieve relevant, valuable memories, particularly those containing successful actions for similar past goals, as highlighted in~\cite{wang2023survey}. This process aids in enhancing their current actions.  
\textbf{2) Self-Evolution.}  Instead of  {only relying on the historical records to decide subsequent actions as seen in  \emph{Memory}-based solutions, agents can dynamically self-evolve by modifying themselves such as altering their initial goals and planning strategies, and training themselves based on feedback or communication logs. \cite{nascimento2023self} proposes a self-control loop process to allow each agent in the multi-agents systems to be self-managed and self-adaptive to dynamic environments, thereby improving the cooperation efficiency of multiple agents. {
\cite{zhang2023proagent} introduces ProAgent which anticipates teammates' decisions and dynamically adjusts each agent's strategies based on the communication logs between agents, facilitating mutual understanding and improving collaborative planning capability.} \cite{wang2023adapting} discusses a  \emph{Learning through Communication} (LTC) paradigm, using the communication logs of multi-agents to generate datasets to train or fine-tune LLMs. LTC enables continuous adaptation and improvement of agents through interaction with their environments and other agents, breaking the limits of in-context learning or supervised fine-tuning, which don't fully utilize the feedback received during interactions with the environment and external tools for continuous training. Self-Evolution enables agents' autonomous adjustment in their profiles or goals, rather than just learning from historical interactions.} 
\textbf{3) Dynamic Generation.} In some scenarios,   the system can generate new agents on-the-fly during its operation~\cite{chen2023autoagents,chen2023agentverse}. This capability enables the system to scale and adapt effectively, as it can introduce agents that are specifically designed to address current needs and challenges. 

{With the scaling up LLM-MA with  a larger  number of agents,  the escalating complexity of managing various kinds of agents has been a critical problem. \emph{Agents Orchestration} emerged as a pivotal challenge and began to gain attention in ~\cite{multiorches1,multiorches2}. We will further discuss this topic in Section \ref{scale_up}.}

\begin{table*}[h!]
\centering
\scalebox{0.59}{  
\begin{tabular}{c|cc|c|c|cc|cc|cc}
\toprule
 & &    &    &  {\color[HTML]{BF9000} } &
  \multicolumn{2}{c|}{{\color[HTML]{4472C4} \textbf{ Agents   Profiling }}} &
  \multicolumn{2}{c|}{{\color[HTML]{548135} \textbf{\begin{tabular}[c]{@{}c@{}}Agents \\ Communication\end{tabular}}}} &
  \multicolumn{2}{c}{{\color[HTML]{FF00FF} \textbf{Agents Capabilities Acquisition}}} \\ \cline{6-11} 
\multirow{-2}{*}{\textbf{Motivation}} &
  \multicolumn{2}{c|}{\multirow{-2}{*}{\textbf{Research Domain \& Goals}}} &
  \multirow{-2}{*}{\textbf{Work}} &
  \multirow{-4}{*}{{\color[HTML]{BF9000} \textbf{\begin{tabular}[c]{@{}c@{}}Agents-Env.\\ Interface  \end{tabular}}}} &
  \multicolumn{1}{c|}{{\color[HTML]{4472C4} \textbf{ \begin{tabular}[c]{@{}c@{}}Profiling\\ methods\end{tabular}}   }} &  
  {\color[HTML]{4472C4} \textbf{\begin{tabular}[c]{@{}c@{}}Profiles\\ (examples)\end{tabular}}} &
  \multicolumn{1}{c|}{{\color[HTML]{548135} \textbf{Paradigms}}} &
  {\color[HTML]{548135}  \textbf{Structure}} &
  \multicolumn{1}{c|}{{\color[HTML]{FF00FF} \textbf{Feedback from}}} &
  {\color[HTML]{FF00FF} \textbf{\begin{tabular}[c]{@{}c@{}}Agents \\ Adjustment \end{tabular}}}     \vspace{1pt} \\ \hline 
 &
  \multicolumn{2}{c|}{} &
 {\scriptsize  \cite{qian2023communicative}} &
  Sandbox &
  \multicolumn{1}{c|}{\begin{tabular}[c]{@{}c@{}}Pre-defined, \\ Model-Generated\end{tabular}} &
  \begin{tabular}[c]{@{}c@{}}CTO,\\ programmer\end{tabular} &
  \multicolumn{1}{c|}{Cooperative} &
  Layered &  
  \multicolumn{1}{c|}{   \begin{tabular}[c]{@{}c@{}}Environment, \\ Agent interaction,\\ Human\end{tabular}} &
  \begin{tabular}[c]{@{}c@{}}Memory,\\ Self-Evolution\end{tabular} \\ \cline{4-11} 
 &
  \multicolumn{2}{c|}{Software development} &
  { \scriptsize \cite{hong2023metagpt}} &
  Sandbox &
  \multicolumn{1}{c|}{Pre-defined} &
  \begin{tabular}[c]{@{}c@{}}Product Manager, \\ Engineer\end{tabular} &
  \multicolumn{1}{c|}{Cooperative} &
   \begin{tabular}[c]{@{}c@{}}Layered, \\ Shared Message Pool\end{tabular} &
  \multicolumn{1}{c|}{ \begin{tabular}[c]{@{}c@{}}Environment, \\ Agent interaction,\\ Human\end{tabular}} &
  \begin{tabular}[c]{@{}c@{}}Memory,\\ Self-Evolution\end{tabular} \\ \cline{4-11} 
 &
  \multicolumn{2}{c|}{ } &
  { \scriptsize \cite{dong2023selfcollaboration}} &
  Sandbox &
  \multicolumn{1}{c|}{\begin{tabular}[c]{@{}c@{}}Pre-defined, \\ Model-Generated\end{tabular}} &
  \begin{tabular}[c]{@{}c@{}}Analyst,\\ coder\end{tabular} &
  \multicolumn{1}{c|}{Cooperative} &
  Layered &
  \multicolumn{1}{c|}{\begin{tabular}[c]{@{}c@{}}Environment,\\  Agent interaction\end{tabular}} &
  \begin{tabular}[c]{@{}c@{}}Memory,\\ Self-Evolution\end{tabular} \\ \cline{2-11} 
 &
  \multicolumn{1}{c|}{} & \begin{tabular}[c]{@{}c@{}}Multi-robot  \\ planning \end{tabular} &
 { \hspace{-10pt}  \scriptsize  \cite{chen2023scalable} \hspace{-10pt} } &
  \begin{tabular}[c]{@{}c@{}}Sandbox, \\ Physical\end{tabular} &
  \multicolumn{1}{c|}{Pre-defined} &
  Robots &
  \multicolumn{1}{c|}{Cooperative} &
  \begin{tabular}[c]{@{}c@{}}Centralized,\\ Decentralized\end{tabular} &
  \multicolumn{1}{c|}{\begin{tabular}[c]{@{}c@{}}Environment,\\  Agent interaction\end{tabular}} &
  Memory \\ \cline{3-11} 
 &
    \multicolumn{1}{c|}{\begin{tabular}[c]{@{}c@{}}Embodied\\ Agents \end{tabular} } &
   \begin{tabular}[c]{@{}c@{}}Multi-robot  \\ collaboration \end{tabular} &
 { \scriptsize \cite{mandi2023roco}} &
  \begin{tabular}[c]{@{}c@{}}Sandbox, \\ Physical\end{tabular} &
  \multicolumn{1}{c|}{Pre-defined} &
  Robots &
  \multicolumn{1}{c|}{Cooperative} &
  Decentralized &
  \multicolumn{1}{c|}{\begin{tabular}[c]{@{}c@{}}Environment,\\ Agent interaction\end{tabular}} &
  Memory \\ \cline{3-11} 
 &
  \multicolumn{1}{c|}{} &
 \begin{tabular}[c]{@{}c@{}}Multi-Agents  \\ cooperation \end{tabular} &
 {\hspace{-15pt} \scriptsize \cite{zhang2023building} \hspace{-15pt}} &
  Sandbox &
  \multicolumn{1}{c|}{Pre-defined} &
  Robots &
  \multicolumn{1}{c|}{Cooperative} &
  Decentralized &
  \multicolumn{1}{c|}{\begin{tabular}[c]{@{}c@{}}Environment,\\ Agent interaction\end{tabular}} &
  Memory \\ \cline{2-11} 
 \multirow{-5}{*}{\begin{tabular}[c]{@{}c@{}}Problem \\ Solving\end{tabular}}  &
   \multicolumn{1}{c|}{\begin{tabular}[c]{@{}c@{}} Science \\ Experiments \end{tabular}} &
   \begin{tabular}[c]{@{}c@{}} Optimization \\ of MOF \end{tabular} &
 {\hspace{-10pt} \scriptsize  \cite{chatgptforchem} \hspace{-10pt} } &
  Physical &
  \multicolumn{1}{c|}{Pre-defined} &
  \begin{tabular}[c]{@{}c@{}}Strategy planers,\\ literature \\ collector, coder\end{tabular} &
  \multicolumn{1}{c|}{Cooperative} &
  Centralized &
  \multicolumn{1}{c|}{\begin{tabular}[c]{@{}c@{}}Environment,\\ Human\end{tabular}} &
  Memory \\ \cline{2-11} 
 &
  \multicolumn{1}{c|}{} &
   \begin{tabular}[c]{@{}c@{}} Improving \\ Factuality \end{tabular} &
 { \scriptsize \cite{du2023improving}} &
  None &
  \multicolumn{1}{c|}{Pre-defined} &
  Agents &
  \multicolumn{1}{c|}{Debate} &
  Decentralized &
  \multicolumn{1}{c|}{Agent interaction} &
  Memory \\ \cline{3-11} 
 &
    \multicolumn{1}{c|}{\begin{tabular}[c]{@{}c@{}}Science \\ Debate\end{tabular}} &
   \begin{tabular}[c]{@{}c@{}}Examining,\\ Inter-Consistency\end{tabular} &
 {\hspace{-15pt} \scriptsize \cite{xiong2023examining} \hspace{-10pt}} &
  None &
  \multicolumn{1}{c|}{Pre-defined} &
  \begin{tabular}[c]{@{}c@{}}Proponent,\\ Opponent,\\ Judge\end{tabular} &
  \multicolumn{1}{c|}{Debate} &
  \begin{tabular}[c]{@{}c@{}}Centralized,\\ Decentralized\end{tabular} &
  \multicolumn{1}{c|}{Agent interaction} &
  Memory \\ \cline{3-11} 
 &
  \multicolumn{1}{c|}{} &
  \begin{tabular}[c]{@{}c@{}}Evaluators \\ for debates \end{tabular} &
 { \hspace{-10pt} \scriptsize \cite{chan2023chateval}} &
  None &
  \multicolumn{1}{c|}{Pre-defined} &
  Agents &
  \multicolumn{1}{c|}{Debate} &
  \begin{tabular}[c]{@{}c@{}}Centralized,\\ Decentralized\end{tabular} &
  \multicolumn{1}{c|}{Agent interaction} &
  Memory \\ \cline{3-11} 
&
    \multicolumn{1}{c|}{} &
    \begin{tabular}[c]{@{}c@{}}Multi-Agents \\ for Medication\end{tabular} &
 {\hspace{-10pt}  \scriptsize \cite{tang2023medagents} \hspace{-10pt} } &
  None &
  \multicolumn{1}{c|}{Pre-defined} &
  \begin{tabular}[c]{@{}c@{}}Cardiology,\\ Surgery\end{tabular} &
  \multicolumn{1}{c|}{\begin{tabular}[c]{@{}c@{}}Debate,\\ Cooperative\end{tabular}} &
  \begin{tabular}[c]{@{}c@{}}Centralized,\\ Decentralized\end{tabular} &
  \multicolumn{1}{c|}{Agent interaction} &
  Memory \\ \hline
 &
  \multicolumn{1}{c|}{} &
    \begin{tabular}[c]{@{}c@{}} Modest Community \\ (25 persons)\end{tabular} &
 {\hspace{-10pt}  \scriptsize \cite{park2023generative} \hspace{-10pt} } &
  Sandbox &
  \multicolumn{1}{c|}{Model-generated} &
  \begin{tabular}[c]{@{}c@{}}Pharmacy,\\ shopkeeper\end{tabular} &
  \multicolumn{1}{c|}{-} &
  - &
  \multicolumn{1}{c|}{\begin{tabular}[c]{@{}c@{}}Environment,\\ Agent interaction\end{tabular}} &
  Memory \\ \cline{3-11} 
 &
  \multicolumn{1}{c|}{} &
  \begin{tabular}[c]{@{}c@{}}  Online community \\ (1000 persons)\end{tabular} &
 {\hspace{-10pt}  \scriptsize \cite{park2022social} \hspace{-10pt} } &
  None &
  \multicolumn{1}{c|}{\begin{tabular}[c]{@{}c@{}}Pre-defined,\\ Model-generated\end{tabular}} &
  \begin{tabular}[c]{@{}c@{}}Camping,\\  fishing\end{tabular} &
  \multicolumn{1}{c|}{-} &
  - &
  \multicolumn{1}{c|}{Agent interaction} &
 \begin{tabular}[c]{@{}c@{}}Dynamic\\  Generation\end{tabular}    \\ \cline{3-11} 
 &
  \multicolumn{1}{c|}{Society} &
  Emotion propagation &
 {\hspace{-10pt}  \scriptsize \cite{gao2023s} \hspace{-10pt} } &
  None &
  \multicolumn{1}{c|}{\begin{tabular}[c]{@{}c@{}}Pre-defined,\\ Model-generated\end{tabular}} &
  \begin{tabular}[c]{@{}c@{}}Real-world \\ user \end{tabular} &
  \multicolumn{1}{c|}{-} &
  - &
  \multicolumn{1}{c|}{Agent interaction} &
  Memory \\ \cline{3-11} 
 &
  \multicolumn{1}{c|}{} &
    \begin{tabular}[c]{@{}c@{}}  Real-time \\ social interactions \end{tabular} &
 {\hspace{-15pt} \scriptsize \cite{kaiya2023lyfe} \hspace{-15pt}} &
  Sandbox &
  \multicolumn{1}{c|}{Pre-defined} &
  \begin{tabular}[c]{@{}c@{}}Real-world \\ user \end{tabular} &
  \multicolumn{1}{c|}{-} &
  - &
  \multicolumn{1}{c|}{\begin{tabular}[c]{@{}c@{}}Environment,\\ Agent interaction\end{tabular}} &
  Memory \\ \cline{3-11} 
 &
  \multicolumn{1}{c|}{ } &
  Opinion dynamics &
 {\hspace{-10pt}  \scriptsize \cite{li2023quantifying}} &
  None &
  \multicolumn{1}{c|}{Pre-defined} &
  \begin{tabular}[c]{@{}c@{}}NIN, NINL,\\ NIL\end{tabular} &
  \multicolumn{1}{c|}{-} &
  - &
  \multicolumn{1}{c|}{Agent interaction} &
  Memory \\ \cline{2-11} 
 &
  \multicolumn{1}{c|}{} &
  WereWolf &
  \begin{tabular}[c]{@{}c@{}} {\scriptsize \cite{xu2023exploring} }\\ {\scriptsize  \cite{xu2023language} }\end{tabular} &
  Sandbox &
  \multicolumn{1}{c|}{Pre-defined} &
  \begin{tabular}[c]{@{}c@{}}Seer,\\ werewolf,\\ villager\end{tabular} &
  \multicolumn{1}{c|}{\begin{tabular}[c]{@{}c@{}}Cooperative, \\ Debate,\\ Competitive\end{tabular}} &
  Decentralized &
  \multicolumn{1}{c|}{\begin{tabular}[c]{@{}c@{}}Environment,\\ Agent interaction\end{tabular}} &
  Memory \\ \cline{3-11} 
 &
  \multicolumn{1}{c|}{Gaming} &
  Avalon &
  \begin{tabular}[c]{@{}c@{}} {\hspace{-10pt} \scriptsize \cite{light2023avalonbench} \hspace{-10pt} } \\ {\hspace{-10pt} \scriptsize \cite{wang2023avalon} \hspace{-10pt} } \end{tabular} &
  Sandbox &
  \multicolumn{1}{c|}{Pre-defined} &
  \begin{tabular}[c]{@{}c@{}}Servant,\\ Merlin,\\  Assassin\end{tabular} &
  \multicolumn{1}{c|}{\begin{tabular}[c]{@{}c@{}}Cooperative, \\ Debate,\\ Competitive\end{tabular}} &
  Decentralized &
  \multicolumn{1}{c|}{\begin{tabular}[c]{@{}c@{}}Environment,\\ Agent interaction\end{tabular}} &
  Memory \\ \cline{3-11} 
 &
  \multicolumn{1}{c|}{} &
  Welfare Diplomacy &
 {\hspace{-10pt}  \scriptsize \cite{mukobi2023welfare} \hspace{-10pt}} &
  Sandbox &
  \multicolumn{1}{c|}{Pre-defined} &
  Countries &
  \multicolumn{1}{c|}{\begin{tabular}[c]{@{}c@{}}Cooperative,\\ Competitive\end{tabular}} &
  Decentralized &
  \multicolumn{1}{c|}{\begin{tabular}[c]{@{}c@{}}Environment,\\ Agent interaction\end{tabular}} &
  Memory \\ \cline{2-11} 
 &
  \multicolumn{1}{c|}{} &
    \begin{tabular}[c]{@{}c@{}}  Human behavior \\ Simulation\end{tabular} &
{\hspace{-10pt}  \scriptsize  \cite{aher2023using} } &
  Sandbox &
  \multicolumn{1}{c|}{Pre-defined} &
  Humans &
  \multicolumn{1}{c|}{-} &
  - &
  \multicolumn{1}{c|}{Agent interaction} &
  Memory \\ \cline{3-11} 
\multirow{-18}{*}{\begin{tabular}[c]{@{}c@{}}World \\ \hspace{-10pt} Simulation \hspace{-10pt} \end{tabular}} 
 &
  \multicolumn{1}{c|}{\multirow{-2}{*}{Psychology}} &
  \begin{tabular}[c]{@{}c@{}}Collaboration\\ Exploring \end{tabular} &
  {\hspace{-10pt} \scriptsize \cite{zhang2023exploring} \hspace{-10pt} } &
  None &
  \multicolumn{1}{c|}{Pre-defined} &
  Agents &
  \multicolumn{1}{c|}{\begin{tabular}[c]{@{}c@{}}Cooperative,\\ Debate\end{tabular}} &
  Decentralized &
  \multicolumn{1}{c|}{Agent interaction} &
  Memory \\ \cline{2-11} 
 &
  \multicolumn{1}{c|}{} &
    \begin{tabular}[c]{@{}c@{}}Macroeconomic \\ simulation \end{tabular} &
  {\hspace{-10pt}  \scriptsize \cite{li2023large}} &
  None &
  \multicolumn{1}{c|}{\begin{tabular}[c]{@{}c@{}}Pre-defined,\\ Model-generated\end{tabular}} &
  Labor &
  \multicolumn{1}{c|}{Cooperative} &
  Decentralized &
  \multicolumn{1}{c|}{Agent interaction} &
  Memory \\ \cline{3-11} 
 &
  \multicolumn{1}{c|}{Economy} &
  \begin{tabular}[c]{@{}c@{}}Information \\ Marketplaces \end{tabular} &
 {\hspace{-10pt}  \scriptsize \cite{anonymous2023rethinking} \hspace{-10pt} } &
  Sandbox &
  \multicolumn{1}{c|}{\begin{tabular}[c]{@{}c@{}}Pre-defined, \\ Data-Derived\end{tabular}} &
  Buyer &
  \multicolumn{1}{c|}{\begin{tabular}[c]{@{}c@{}}Cooperative, \\ Competitive\end{tabular}} &
  Decentralized &
  \multicolumn{1}{c|}{\begin{tabular}[c]{@{}c@{}}Environment,\\ Agent interaction\end{tabular}} &
  Memory \\ \cline{3-11} 
 &
  \multicolumn{1}{c|}{} &
    \begin{tabular}[c]{@{}c@{}}Improving \\ financial trading \end{tabular} &

  {\hspace{-10pt}  \scriptsize \cite{li2023tradinggpt}} &
  Physical &
  \multicolumn{1}{c|}{Pre-defined} &
  Trader &
  \multicolumn{1}{c|}{Debate} &
  Decentralized &
  \multicolumn{1}{c|}{\begin{tabular}[c]{@{}c@{}}Environment,\\ Agent interaction\end{tabular}} &
  Memory \\ \cline{3-11} 
 &
  \multicolumn{1}{c|}{} &
  Economic theories &
  {\hspace{-10pt}  \scriptsize \cite{zhao2023competeai}} &
  Sandbox &
  \multicolumn{1}{c|}{\begin{tabular}[c]{@{}c@{}}Pre-defined, \\ Model-Generated\end{tabular}} &
  \begin{tabular}[c]{@{}c@{}}Restaurant,\\  Customer\end{tabular} &
  \multicolumn{1}{c|}{Competitive} &
  Decentralized &
  \multicolumn{1}{c|}{\begin{tabular}[c]{@{}c@{}}Environment,\\ Agent interaction\end{tabular}} &
  \begin{tabular}[c]{@{}c@{}}Memory,\\ Self-Evolution\end{tabular} \\ \cline{2-11} 
 &
  \multicolumn{1}{c|}{\multirow{2}{*}{ \begin{tabular}[c]{@{}c@{}} \hspace{-7pt} Recommender \hspace{-5pt}  \\ Systems \end{tabular}}} &
  \begin{tabular}[c]{@{}c@{}}Simulating \\ user behaviors\end{tabular} &
  {\hspace{-15pt} \scriptsize \cite{zhang2023generative} \hspace{-10pt}} &
  Sandbox &
  \multicolumn{1}{c|}{Data-Derived} &
  \begin{tabular}[c]{@{}c@{}}Users  from \\ MovieLens-1M\end{tabular} &
  \multicolumn{1}{c|}{-} &
  - &
  \multicolumn{1}{c|}{Environment} &
  Memory \\ \cline{3-11} 
 &
  \multicolumn{1}{c|}{ } &
    \begin{tabular}[c]{@{}c@{}}\hspace{-10pt} Simulating user-item \hspace{-10pt} \\ interactions \end{tabular} &
  { \hspace{-15pt} \scriptsize \cite{zhang2023agentcf} \hspace{-10pt}} &
  Sandbox &
  \multicolumn{1}{c|}{\begin{tabular}[c]{@{}c@{}}Pre-defined, \\ Data-Derived\end{tabular}} &
  \begin{tabular}[c]{@{}c@{}}User Agents\\ Item Agents\end{tabular} &
  \multicolumn{1}{c|}{Cooperative} &
  Decentralized &
  \multicolumn{1}{c|}{\begin{tabular}[c]{@{}c@{}}Environment,\\ Agent interaction\end{tabular}} &
  Memory \\ \cline{2-11} 
 &
  \multicolumn{1}{c|}{\multirow{2}{*}{ \begin{tabular}[c]{@{}c@{}} Policy \\ Making \end{tabular}}} &
  \begin{tabular}[c]{@{}c@{}}Public \\ Administration \end{tabular} &
  {\hspace{-10pt}  \scriptsize \cite{xiao2023simulating}} &
  None &
  \multicolumn{1}{c|}{Pre-defined} &
  Residents &
  \multicolumn{1}{c|}{Cooperative} &
  Decentralized &
  \multicolumn{1}{c|}{Agent interaction} &
  Memory \\ \cline{3-11}  
 &
  \multicolumn{1}{c|}{} &
    War Simulation &
  {\hspace{-10pt}  \scriptsize \cite{hua2023war}} &
  None &
  \multicolumn{1}{c|}{Pre-defined} &
  Countries &
  \multicolumn{1}{c|}{Competitive} &
  Decentralized &
  \multicolumn{1}{c|}{Agent interaction} &
  Memory \\ \cline{2-11} 
  & \multicolumn{1}{c|}{\multirow{2}{*}{ \begin{tabular}[c]{@{}c@{}} Disease \end{tabular}}} &
   \begin{tabular}[c]{@{}c@{}}Human Behaviors  \\ to epidemics\end{tabular} &
 {\hspace{-6pt} \scriptsize \begin{tabular}[c]{@{}c@{}} [Ghaffarzadegan \\ \emph{et al.}, 2023] \end{tabular}}    &
  Sandbox &
  \multicolumn{1}{c|}{\begin{tabular}[c]{@{}c@{}}Pre-defined, \\ Model-Generated\end{tabular}} &
 \begin{tabular}[c]{@{}c@{}}Conformity   \\ traits \end{tabular} &
  \multicolumn{1}{c|}{Cooperative} &
  Decentralized &
  \multicolumn{1}{c|}{\begin{tabular}[c]{@{}c@{}}Environment,\\ Agent interaction\end{tabular}} &
  Memory  \\ \cline{3-11}  \vspace{1pt} 
  & \multicolumn{1}{c|}{} &
  Public health &
  { \scriptsize \begin{tabular}[c]{@{}c@{}} [Williams \\ \emph{et al.}, 2023] \end{tabular}} &
  Sandbox &
  \multicolumn{1}{c|}{\begin{tabular}[c]{@{}c@{}}Pre-defined, \\ Model-Generated\end{tabular}} &
  \begin{tabular}[c]{@{}c@{}} Adults aged  \\ 18 to 64\end{tabular} &
  \multicolumn{1}{c|}{Cooperative} &
  Decentralized &
  \multicolumn{1}{c|}{\begin{tabular}[c]{@{}c@{}}Environment,\\ Agent interaction\end{tabular}} &
  \begin{tabular}[c]{@{}c@{}}Memory,\\ Dynamic \\ Generation\end{tabular} \\ \bottomrule
\end{tabular}
}
\caption{Summary of the LLM-MA studies. We categorize current work according to their motivation, research domains and goals, and detail each work from different aspects regarding 
Agents-Environment Interface, Agents Profiling, Agents Communication and Agents Capability Acquisition. 
``-" denotes that a particular element is not specifically mentioned in this work.}
\label{summary}
\end{table*}

\section{Applications} 
\label{Sec5:Application}


LLM-MA systems have been used in a wide range of applications. We summarize two kinds of applications in Table \ref{summary}: \textbf{Problem Solving} and  \textbf{World Simulation}. We   elaborate on these applications below. Note that this is a fast growing research field and new applications appear almost everyday. We maintain an  \href{https://github.com/taichengguo/LLM_MultiAgents_Survey_Papers}{open source repository} to report the latest work.  

\subsection{LLM-MA for Problem Solving}

The main motivation of using LLM-MA for problem solving is to harness the collective capabilities of agents with specialized expertise. These agents, each acting as individuals, collaborate to address complex problems effectively,  such as software development, embodied agents, science experiments and science debate. These application examples are introduced next.

\subsubsection{Software Development}
Given that software development is a complex endeavor requiring the collaboration of various roles like product managers, programmers, and testers, LLM-MA systems are typically set to emulate these distinct roles and collaborate to address the intricate challenge. Following the waterfall or Standardized Operating Procedures (SOPs) workflow of the software development, the communication structure among agents is usually layered. Agents generally interact with the code interpreter, other agents or human to iteratively refine the generated code.
\cite{li2023camel} first proposes a simple role-play agent framework, which utilizes the interplay of two roles to realize autonomous programming based on one-sentence user instruction. It provides insights into the ``cognitive" processes of communicative agents. \cite{dong2023selfcollaboration} makes LLMs work as distinct ``experts" for sub-tasks in software development, autonomously collaborating to generate code. Moreover, \cite{qian2023communicative} presents an end-to-end framework for software development, utilizing multiple agents for software development without incorporating advanced human teamwork experience. \cite{hong2023metagpt} first incorporates human workflow insights for more controlled and validated performance. It encodes SOPs into prompts to enhance structured coordination. \cite{huang2023agentcoder} delves deeper into multi-agent based programming by solving the problem of balancing code snippet generation with effective test case generation, execution, and optimization. 


\subsubsection{Embodied Agents}
Most embodied agents applications inherently utilize multiple robots working together to perform complex real-world planning and manipulation tasks such as warehouse management with heterogeneous robot capabilities. Hence, LLM-MA can be used to model robots with different capabilities and cooperate with each other to solve real-world physical tasks.
\cite{dasgupta2023collaborating} first explores the potential to use LLM as an action planner for embedded agents. \cite{mandi2023roco} introduces RoCo, a novel approach for multi-robot collaboration that uses LLMs for high-level communication and low-level path planning. Each robotic arm is equipped with an LLM, cooperating with inverse kinematics and collision checking. Experimental results demonstrate the adaptability and success of RoCo in collaborative tasks. \cite{zhang2023building} presents CoELA, a Cooperative Embodied Language Agent, managing discussions and task planning in an LLM-MA setting. This challenging setting is featured with decentralized control, complex partial observation, costly communication, and multi-objective long-horizon tasks. \cite{chen2023scalable} investigates communication challenges in scenarios involving a large number of robots, as assigning each robot an LLM will be costly and unpractical due to the long context. The study compares four communication frameworks, centralized, decentralized, and two hybrid models, to evaluate their effectiveness in coordinating complex multi-agent tasks. \cite{yu2023conavgpt} proposes Co-NavGPT for multi-robot cooperative visual target navigation, integrating LLM as a global planner to assign frontier goals to each robot. \cite{chen2023multi} proposes an LLM-based consensus-seeking framework, which can be applied as a cooperative planner to a multi-robot aggregation task.


\subsubsection{Science Experiments}
Like multiple agents play as different specialists and cooperate to solve the Software Development and Embodied Agents problem, multiple agents can also be used to form a science team to conduct science experiments. One important difference from previous applications lies in the crucial role of human oversight,  due to the high expenses of the science experiments and the hallucination of the LLM agents. Human experts are at the center of these agents to process the information of agents and give feedback to the agents.
~\cite{chatgptforchem} utilizes multiple LLM-based agents, each focusing on specific tasks for the science experiments including strategy planning, literature search, coding, robotic operations, and labware design. All these agents interact with humans to work collaboratively to optimize the synthesis process of complex materials.

\subsubsection{Science Debate}
LLM-MA can be set for science debating scenarios, where agents debate with each other to enhance the collective reasoning capabilities in tasks such as Massive Multitask Language Understanding (MMLU)~\cite{hendrycks2020measuring}, Math problems~\cite{cobbe2021training}, and StrategyQA~\cite{geva2021did}. The main idea is that each agent initially offers its own analysis of a problem, which is then followed by a joint debating process. Through multiple rounds of debate, the agents converge on a single, consensus answer. 
~\cite{du2023improving} leverages the multi-agents debate process on a set of six different reasoning and factual accuracy tasks and demonstrates that LLM-MA debating can improve factuality. ~\cite{xiong2023examining} focuses on the commonsense reasoning tasks and formulates a three-stage debate to align with real-world scenarios including fair debate, mismatched debate, and roundtable debate. The paper also analyzes the inter-consistency between different LLMs and claims that debating can improve the inter-consistency.~\cite{tang2023medagents} also utilizes multiple LLM-based agents as distinct domain experts to do the collaborative discussion on a medical report to reach a consensus for medical diagnosis.

\subsection{LLM-MA for World Simulation}

Another mainstream application scenario of LLM-MA is the world simulation.  Research in this area is rapidly growing and spans a diverse range of fields including social sciences, gaming, psychology, economics, policy-making, etc.
The key reason for employing LLM-MA in world simulations lies in their exceptional role-playing abilities, which are crucial for realistically depicting various roles and viewpoints in a simulated world.
The environment of world simulation projects is usually crafted to reflect the specific scenario being simulated, with agents designed in various profiles to match this context. Unlike the problem solving systems that focus on agent cooperation, world simulation systems involve diverse methods of agent management and communication, reflecting the complexity and variety of real-world interactions.
Next, we explore simulations conducted in diverse fields.

\subsubsection{Societal Simulation}
In societal simulation, LLM-MA models are used to simulate social behaviors, aiming to explore the potential social dynamics and propagation, test social science theories, and populate virtual spaces and communities with realistic social phenomena~\cite{park2023generative}.  Leveraging LLM's capabilities, agents with unique profiles engage in extensive communication, generating rich behavioral data for in-depth social science analysis. 

The scale of societal simulation has expanded over time, beginning with smaller, more intimate settings and progressing to larger, more intricate ones. 
Initial work by \cite{park2023generative} introduces generative agents within an interactive sandbox environment reminiscent of the sims, allowing end users to engage with a modest community of 25 agents through natural language. 
At the same time, \cite{park2022social} develops Social Simulacra, which constructs a simulated community of 1,000 personas. This system takes a designer's vision for a community—its goals, rules, and member personas—and simulates it, generating behaviors like posting, replying, and even anti-social actions.
Building on this, \cite{gao2023s} takes the concept further by constructing vast networks comprising 8,563 and 17,945 agents, respectively, designed to simulate social networks focused on the topics of Gender Discrimination and Nuclear Energy. This evolution showcases the increasing complexity and size of simulated environments in recent research.
Recent studies such as \cite{chen2023multi,kaiya2023lyfe,li2023quantifying,li2023you,ziems2023can} highlight the evolving complexity in multi-agent systems, LLM impacts on social networks, and their integration into social science research.
 
\subsubsection{Gaming}
LLM-MA is well-suited for creating simulated gaming environments, allowing agents to assume various roles within games. This technology enables the development of controlled, scalable, and dynamic settings that closely mimic human interactions, making it ideal for testing a range of game theory hypotheses~\cite{mao2023alympics,xu2023exploring,gong2023mindagent}. 
Most games simulated by LLM-MA rely heavily on natural language communication, offering a sandbox environment within different game settings for exploring or testing game theory hypotheses including reasoning, cooperation, persuasion, deception, leadership, etc. 

~\cite{akata2023playing} leverages behavioral game theory to examine LLMs' behavior in interactive social settings, particularly their performance in games like the iterated Prisoner's Dilemma and Battle of the Sexes. Furthermore, \cite{xu2023exploring} proposes a framework using ChatArena library \cite{ChatArena} for engaging LLMs in communication games like Werewolf, using retrieval and reflection on past communications for improvement, as well as the Chain-of-Thought mechanism \cite{wei2022chain}.
~\cite{light2023text} explores the potential of LLM agents in playing Resistance Avalon, introducing AVALONBENCH, a comprehensive game environment and benchmark for further developing advanced LLMs and multi-agent frameworks. ~\cite{wang2023avalon} also focuses on the capabilities of LLM Agents in dealing with misinformation in the Avalon game, proposing the Recursive Contemplation (ReCon) framework to enhance LLMs' ability to discern and counteract deceptive information. 
~\cite{xu2023language} introduces a framework combining LLMs with reinforcement learning (RL) to develop strategic language agents for the Werewolf game. It introduces a new approach to use RL policy in the case that the action and state sets are not predefined but in the natural language setting. ~\cite{mukobi2023welfare} designs the ``Welfare Diplomacy”, a general-sum variant of the zero-sum board game Diplomacy, where players must balance military conquest and domestic welfare. It also offers an open-source benchmark, aiming to help improve the cooperation ability of multi-agent AI systems. On top of that, there is a work \cite{li2023theory} in a multi-agent cooperative text game testing the agents' Theory of Mind (ToM), the ability to reason about the concealed mental states of others and is fundamental to human social interactions, collaborations, and communications. \cite{fan2023can} comprehensively assesses the capability of LLMs as rational players, and identifies the weaknesses of LLM-based Agents that even in the explicit game process, agents may still overlook or modify refined beliefs when taking actions.

\subsubsection{Psychology}
In psychological simulation studies, like in the societal simulation, multiple agents are utilized to simulate humans with various traits and thought processes. However, unlike societal simulations, one approach in psychology involves directly applying psychological experiments to these agents. This method focuses on observing and analyzing their varied behaviors through statistical methods. Here, each agent operates independently, without interacting with others, essentially representing different individuals. Another approach aligns more closely with societal simulations, where multiple agents interact and communicate with each other. In this scenario, psychological theories are applied to understand and analyze the emergent behavioral patterns. This method facilitates the study of interpersonal dynamics and group behaviors, providing insights into how individual psychological traits influence collective actions.
~\cite{ma2023understanding} explores the psychological implications and outcomes of employing LLM-based conversational agents for mental well-being support. It emphasizes the need for carefully evaluating the use of LLM-based agents in mental health applications from a psychological perspective. 
~\cite{kovavc2023socialai} introduces a tool named SocialAI school for creating interactive environments simulating social interactions. It draws from developmental psychology to understand how agents can acquire, demonstrate, and evolve social skills such as joint attention, communication, and cultural learning. 
~\cite{zhang2023exploring} explores how LLM agents, with distinct traits and thinking patterns, emulate human-like social behaviors such as conformity and majority rule. This integration of psychology into the understanding of agent collaboration offers a novel lens for examining and enhancing the mechanisms behind LLM-based multi-agents systems.
~\cite{aher2023using} introduces Turing Experiments to evaluate the extent to which large language models can simulate different aspects of human behaviors. The Turing Experiments replicate classical experiments and phenomena in psychology, economics, and sociology using a question-answering format to mimic experimental conditions. They also design a prompt that is used to simulate the responses of multiple
different individuals by varying the name. By simulating various kinds of individuals via LLM, they show that larger models replicate human behavior more faithfully, but they also reveal a hyper-accuracy distortion, especially in knowledge-based tasks.

\begin{table*}[ht]
\centering
\resizebox{0.9\textwidth}{!}{
\begin{tabular}{l|l|l|l|l}
\hline
\textbf{Motivation} &
  \textbf{Domain} &
  \textbf{Datasets and Benchmarks} &
  \textbf{Used by} &
  \textbf{Data Link} \\ \hline
\multirow{13}{*}{Problem Solving} &
  \multirow{3}{*}{Software Development} &
  HumanEval &
  ~\cite{hong2023metagpt} & \href{https://github.com/openai/human-eval}{Link} \\
 &
   &
  MBPP &
  ~\cite{hong2023metagpt} & \href{https://github.com/google-research/google-research/tree/master/mbpp}{Link}
   \\
 &
   &
  SoftwareDev &
  ~\cite{hong2023metagpt} &
  \href{https://github.com/geekan/MetaGPT}{Link} \\ \cline{2-5} 
 &
  \multirow{4}{*}{Embodied AI} &
  RoCoBench &
  ~\cite{mandi2023roco} &
  \href{https://github.com/MandiZhao/robot-collab/tree/main/rocobench}{Link} \\
 &
   &
  Communicative Watch-And-Help (C-WAH) &
  ~\cite{zhang2023building} &
  \href{https://github.com/UMass-Foundation-Model/Co-LLM-Agents/}{Link} \\
 &
   &
  ThreeDWorld Multi-Agent Transport (TDW-MAT) &
  ~\cite{zhang2023building} &
  \href{https://github.com/UMass-Foundation-Model/Co-LLM-Agents/}{Link} \\
 &
   &
  HM3D v0.2 &
  ~\cite{yu2023conavgpt} &
  \href{https://aihabitat.org/datasets/hm3d-semantics/}{Link} \\ \cline{2-5} 
 &
  \multirow{6}{*}{Science Debate} &
  MMLU &
  ~\cite{tang2023medagents} &
  \href{https://github.com/hendrycks/test}{Link} \\
 &
   &
  MedQA &
   ~\cite{tang2023medagents} &
  \href{https://github.com/jind11/MedQA}{Link} \\
 &
   &
  PubMedQA &
  ~\cite{tang2023medagents} &
  \href{https://github.com/pubmedqa/pubmedqa}{Link} \\
 &
   &
  GSM8K &
  ~\cite{du2023improving} &
  \href{https://github.com/openai/grade-school-math}{Link} \\
 &
   &
  StrategyQA &
  ~\cite{xiong2023examining} &
  \href{https://github.com/eladsegal/strategyqa}{Link} \\
 &
   &
  Chess Move Validity &
  ~\cite{du2023improving} &
  \href{https://github.com/google/BIG-bench}{Link} \\ \hline
\multirow{15}{*}{World Simulation} &
  \multirow{3}{*}{Society} &
  SOTOPIA &
  ~\cite{zhou2023sotopia} &
  / \\
 &
   &
  Gender Discrimination &
  ~\cite{gao2023s} &
  / \\
 &
   &
  Nuclear Energy &
  ~\cite{gao2023s} &
  / \\ \cline{2-5} 
 &
  \multirow{6}{*}{Gaming} &
  Werewolf &
  ~\cite{xu2023exploring} &
  / \\
 &
   &
  Avalon &
  ~\cite{light2023text} &
  / \\
 &
   &
  Welfare Diplomacy &
  ~\cite{mukobi2023welfare} &
  / \\
 &
   &
  Layout  in the Overcooked-AI environment &
  ~\cite{agashe2023evaluating} &
  / \\
 &
   &
  Chameleon &
  ~\cite{xu2023magic} &
  \href{https://github.com/cathyxl/MAgIC}{Link} \\
 &
   &
  Undercover &
  ~\cite{xu2023magic} &
  \href{https://github.com/cathyxl/MAgIC}{Link} \\ \cline{2-5} 
 &
  \multirow{3}{*}{Psychology} &
  Ultimatum Game TE &
  ~\cite{aher2023using} &
  \href{https://github.com/microsoft/turing-experiments/tree/main}{Link} \\
 &
   &
  Garden Path TE &
  ~\cite{aher2023using} &
  \href{https://github.com/microsoft/turing-experiments/tree/main}{Link} \\
 &
   &
  Wisdom of Crowds TE &
  ~\cite{aher2023using} &
  \href{https://github.com/microsoft/turing-experiments/tree/main}{Link} \\ \cline{2-5} 
 &
  \multirow{2}{*}{Recommender System} &
  MovieLens-1M &
  ~\cite{zhang2023generative} &
  \href{https://github.com/LehengTHU/Agent4Rec/tree/master/datasets/ml-1m}{Link} \\
 &
   &
  Amazon review dataset &
  ~\cite{zhang2023agentcf} &
  / \\ \cline{2-5} 
 &
  Policy Making &
  Board Connectivity Evaluation &
  ~\cite{hua2023war} &
  \href{https://github.com/agiresearch/WarAgent}{Link} \\ \hline
\end{tabular}
}
\caption{Datasets and Benchmarks commonly used in LLM-MA studies. `` / " denotes the unavailability of  data link.}
\label{db}
\end{table*}

\subsubsection{Economy}
LLM-MA is used to simulate economic and financial trading environments mainly because it can serve as implicit computational models of humans. In these simulations, agents are provided with endowments, and information, and set with pre-defined preferences, allowing for an exploration of their actions in economic and financial contexts. This is similar to the way economists model 'homo economicus', the characterization of man in some economic theories as a rational person who pursues wealth for his own self-interest ~\cite{horton2023large}. 
There are several studies demonstrate the diverse applications of LLM-MA in simulating economic scenarios, encompassing macroeconomic activities, information marketplaces, financial trading, and virtual town simulations. Agents interact in cooperative or debate, decentralized environments. \cite{li2023large} employs LLMs for macroeconomic simulation, featuring prompt-engineering-driven agents that emulate human-like decision-making, thereby enhancing the realism of economic simulations compared to rule-based or other AI agents. \cite{anonymous2023rethinking} explores the buyer's inspection paradox in an information marketplace, reveals improved decision-making and answer quality when agents temporarily access information before purchase. \cite{li2023tradinggpt} presents an LLM-MA framework for financial trading, emphasizing a layered memory system, debate mechanisms, and individualized trading characters, thereby fortifying decision-making robustness. \cite{zhao2023competeai} utilizes LLM-based Agents to simulate a virtual town with restaurant and customer agents, yielding insights aligned with sociological and economic theories. These studies collectively illuminate the broad spectrum of applications and advancements in employing LLMs for diverse economic simulation scenarios.

\subsubsection{Recommender Systems}
The use of the LLM-MA in recommender systems is similar to that in psychology since studies in both fields involve the consideration of extrinsic and intrinsic human factors such as cognitive processes and personality~\cite{PsychologyInformedRecommenderSystems}. One way to use LLM-MA in recommender systems is to directly introduce items to multiple LLM-based agents within diverse traits and conduct statistics of the preferences of different agents. Another way is to treat both users and items as agents and the user-item communication as interactions, simulating the preference propagation. To bridge the gap between offline metrics and real-world performance in recommendation systems, Agent4Rec~\cite{zhang2023generative} introduces a simulation platform based on LLM-MA. 1000 generative agents are initialized with the MovieLens-1M dataset to simulate complex user interactions in a recommendation environment. 
Agent4Rec shows that LLM-MA can effectively mimic real user preferences and behaviors, provide insights into phenomena like the filter bubble effect, and help uncover causal relationships in recommendation tasks. In Agent4Rec work, agents are used to simulate users and they do not communicate with each other. Different from Agent4Rec work, ~\cite{zhang2023agentcf} treats both users and items as agents, optimizing them collectively to reflect and adjust to real-world interaction disparities. This work emphasizes simulating user-item interactions and propagates preferences among agents, capturing the collaborative filtering essence.

\subsubsection{Policy Making}
Similar to simulations in gaming and economic scenarios, Policy Making requires strong decision-making capabilities to realistic and dynamic complex problems. LLM-MA can be used to simulate the policy making via simulating a virtual government or simulating the impact of various policies on different communities. These simulations provide valuable insights into how policies are formulated and their potential effects, aiding policymakers in understanding and anticipating the consequences of their decisions~\cite{policy}.
The research outlined in  \cite{xiao2023simulating} is centered on simulating a township water pollution crisis. It simulated a town located on an island including a demographic structure of different agents and township head and advisor. Within the water pollution crisis simulation, this work provides an in-depth analysis of how a virtual government entity might respond to such a public administration challenge and how information transfer in the social network in this crisis. ~\cite{hua2023war} introduces WarAgent to simulate key historical conflicts and provides insights for conflict resolution and understanding, with potential applications in preventing future international conflicts.

\subsubsection{Disease Propagation Simulation}
Leveraging the societal simulation capabilities of LLM-MA can also be used to simulate disease propagation.
The most recent study in \cite{williams2023epidemic} delves into the use of LLM-MA in simulating disease spread. 
The research showcases through various simulations how these LLM-based agents can accurately emulate human responses to disease outbreaks, including behaviors like self-quarantine and isolation during heightened case numbers. 
The collective behavior of these agents mirrors the complex patterns of multiple waves typically seen in pandemics, eventually stabilizing into an endemic state. 
Impressively, their actions contribute to the attenuation of the epidemic curve. 
~\cite{ghaffarzadegan2023generative} also discusses the epidemic propagation simulation and decomposes the simulation into two parts: the Mechanistic Model which represents the information or propagation of the virus and the Decision-Making Model which represents the agents' decision-making process when facing the virus.



\section{Implementation Tools and Resources}
\label{Sec4:Framework}
\subsection{Multi-Agents Framework}

We provide a detailed introduction to three open-source multi-agent frameworks: MetaGPT~\cite{hong2023metagpt}, CAMEL~\cite{li2023camel}, and Autogen~\cite{wu2023autogen}.
They are all frameworks that utilize language models for complex task-solving with a focus on multi-agent collaboration, but they differ in their approaches and applications.

MetaGPT is designed to embed human workflow processes into the operation of language model agents, thereby reducing the hallucination problem that often arises in complex tasks. 
It does this by encoding Standard Operating Procedures into the system and using an assembly line approach to assign specific roles to different agents. 

CAMEL, or Communicative Agent Framework, is oriented towards facilitating autonomous cooperation among agents. It uses a novel technique called inception prompting to guide conversational agents towards fulfilling tasks that are consistent with human objectives. This framework also serves as a tool for generating and studying conversational data, helping researchers understand how communicative agents behave and interact.

AutoGen is a versatile framework that allows for the creation of applications using language models. 
It is distinctive for its high level of customization, enabling developers to program agents using both natural language and code to define how these agents interact. 
This versatility enables its use in diverse fields, from technical areas such as coding and mathematics to consumer-focused sectors like entertainment.

More recently, \cite{chen2023agentverse,chen2023autoagents} introduce frameworks for dynamic multi-agent collaboration, while \cite{zhou2023agents,li2023metaagents,xie2023openagents} present platforms and libraries for building autonomous agents, emphasizing their adaptability in task-solving and social simulations.

\subsection{Datasets and Benchmarks} 
\label{Sec6:Datasets and Benchmarks}

We summarize commonly used datasets or benchmarks for LLM-MA study in Table~\ref{db}. We observe that different research applications use different datasets and benchmarks. In the Problem solving scenarios, most datasets and benchmarks are used to evaluate the planning and reasoning capabilities by Multiple agents cooperation or debate. In World Simulation scenarios, datasets and benchmarks are used to evaluate the alignment between the simulated world and real-world or analyze the behaviors of different agents. However, in certain research applications like Science Team operations for experiments and economic modeling, there is still a need for comprehensive benchmarks.   The development of such benchmarks would greatly enhance the ability to gauge the success and applicability of LLM-MA in these complex and dynamic fields.

\section{Challenges and Opportunities}
\label{Sec7:Challenges and Opportunities}
Studies of LLM-MA frameworks and applications are 
advancing rapidly, giving rise to numerous challenges and opportunities.   We identified several critical challenges and potential areas for future study.  

\subsection{Advancing into  Multi-Modal Environment}

Most previous work on LLM-MA has been focused on text-based environments, excelling in processing and generating text. However, there is a notable lack in multi-modal settings, where agents would interact with and interpret data from multiple sensory inputs and generate multiple outputs such as images, audio, video, and physical actions. Integrating LLMs into multi-modal environments presents additional challenges, such as processing diverse data types and enabling agents to understand each other and respond to more than just textual information. 

\subsection{Addressing Hallucination}
The hallucination problem is a significant challenge in LLMs and single LLM-based Agent systems. It refers to the phenomenon where the model generates text that is factually incorrect~\cite{huang2023survey}. However, this problem takes on an added layer of complexity in a multi-agent setting. In such scenarios, one agent's hallucination can have a cascading effect. This is due to the interconnected nature of multi-agent systems, where misinformation from one agent can be accepted and further propagated by others in the network. 
Therefore, detecting and mitigating hallucinations in LLM-MA is not just a crucial task but also presents a unique set of challenges. It involves not only correcting inaccuracies at the level of individual agents but also managing the flow of information between agents to prevent the spread of these inaccuracies throughout the system.


\subsection{Acquiring Collective Intelligence}
In traditional multi-agent systems,  agents often use reinforcement learning to learn from offline training datasets.
However, LLM-MA systems mainly learn from instant feedback, such as interactions with the environment or humans, as we discussed in Section~\ref{Sec3:Taxonomy}.
This learning style requires a reliable interactive environment and it would be tricky to design such an interactive environment for many tasks, limiting the scalability of LLM-MA systems. Moreover, the prevailing approaches in current research involve employing Memory and Self-Evolution techniques to adjust agents based on feedback. While effective for individual agents, these methods do not fully capitalize on the potential collective intelligence of the agent network. They adjust agents in isolation, overlooking the synergistic effects that can emerge from coordinated multi-agent interactions. Hence,  jointly adjusting multiple agents and achieving optimal collective intelligence is still a critical challenge for LLM-MA.


\subsection{Scaling Up LLM-MA Systems}
\label{scale_up}

LLM-MA systems are composed of a number of individual LLM-based agents, posing a significant challenge of scalability regarding the number of agents. 
From the computational complexity perspective, each LLM-based agent, typically built on large language models like GPT-4, demands substantial computational power and memory. Scaling up the number of these agents in an LLM-MA system significantly increases  resource requirements. In scenarios with limited computational resource, it would be challenging to develop these LLM-MA systems.

Additionally, as the number of agents in an LLM-MA system increases, additional complexities and research opportunities emerge, particularly in areas like efficient agent coordination, communication, and understanding the scaling laws of multi-agents. For instance, with more LLM-based agents, the intricacy of ensuring effective coordination and communication rises significantly. 
{As highlighted in ~\cite{multiorches2}, designing advanced Agents Orchestration methodologies is increasingly important. These methodologies aim to optimize agents workflows, task assignments tailored to different agents, and communication patterns across agents such as communication constraints between agents. Effective Agents Orchestration facilitates harmonious operation among agents, minimizing conflicts and redundancies.}
Additionally, exploring and defining the scaling laws that govern the behavior and efficiency of multi-agent systems as they grow larger remains an important area of research. These aspects highlight the need for innovative solutions to optimize LLM-MA systems, making them both effective and resource-efficient.

\subsection{Evaluation and Benchmarks}
We have summarized the datasets and benchmarks currently available for LLM-MA in  Table~\ref{db}. This is a starting point, and far from  being comprehensive. We identify two significant challenges in evaluating  LLM-MA systems and benchmarking their performance against each other. 
Firstly, as discussed in ~\cite{xu2023magic}, much of the existing research focuses on evaluating individual agents' understanding and reasoning within narrowly defined scenarios. This focus tends to overlook the broader and more complex emergent behaviors that are integral to multi-agent systems. 
Secondly, there is a notable shortfall in the development of comprehensive benchmarks across several research domains, such as Science Team for Experiment Operations, Economic analysis, and Disease propagation simulation. This gap presents an obstacle to accurately assessing and benchmarking the full capabilities of LLM-MA systems in these varied and crucial fields.

\subsection{Applications and Beyond}
The potential of LLM-MA systems extends far beyond their current applications, holding great promise for advanced computational problem-solving in fields such as finance, education, healthcare,  environmental science, urban planning and so on. As we have discussed, LLM-MA systems possess the capability to tackle complex problems and simulate various aspects of the real world. While the current role-playing capabilities of LLMs may have limitations, ongoing advancements in LLM technology suggest a bright future. It is anticipated to have more sophisticated methodologies, applications, datasets, and benchmarks tailored for diverse research fields.
Furthermore, there are opportunities to explore LLM-MA systems from various theoretical perspectives, such as Cognitive Science~\cite{sumers2023cognitive}, Symbolic Artificial Intelligence, Cybernetics, Complex Systems, and Collective Intelligence.  Such a multi-faceted approach could  contribute to a more comprehensive understanding and innovative applications in this rapidly evolving field.


\section{Conclusion}
\label{Sec8:Conclusion}
LLM-based Multi-Agents have shown inspiring collective intelligence and rapidly garnered increasing interest among researchers.
In this survey, we first systematically review the development of LLM-MA systems by positioning, differentiating, and connecting them from various aspects, regarding the agents-environment interface, the characterization of agents by LLMs, the strategies for managing agent communication and the paradigms for capability acquisition. We also summarized  LLM-MA applications for problem-solving and world simulation. By also highlighting the commonly used datasets and benchmarks and discussing challenges and future opportunities, we hope that this survey can serve as a useful resource for researchers across various research fields, inspiring future research to explore the potential of LLM-based Multi-Agents.

\appendix


\bibliographystyle{named}
\bibliography{ijcai23}

\end{document}